\documentclass[10pt,journal,compsoc]{IEEEtran}
\usepackage[nocompress]{cite}
\usepackage[pdftex]{graphicx}
\usepackage{amsmath}
\usepackage{algorithmic}
\usepackage{array}
\usepackage[caption=false,font=footnotesize,labelfont=sf,textfont=sf]{subfig}
\usepackage{url}
\usepackage[breaklinks,colorlinks,pagebackref]{hyperref}
\usepackage{ragged2e}
\begin{document}

% ---------- TITLE AND AUTHOR ----------
\title{Multi-scale Attention Guided Pose Transfer}
\author{
  Prasun~Roy,
  Saumik~Bhattacharya,
  Subhankar~Ghosh,
  and~Umapada~Pal%,~\IEEEmembership{Fellow,~IEEE}
  \IEEEcompsocitemizethanks{
    \IEEEcompsocthanksitem P. Roy, S. Ghosh and U. Pal are with the Computer Vision and Pattern Recognition Unit, Indian Statistical Institute, Kolkata, India.%\protect\\
    \IEEEcompsocthanksitem S. Bhattacharya is with Indian Institute of Technology, Kharagpur, India.
    \IEEEcompsocthanksitem Code: \url{https://github.com/prasunroy/pose-transfer}
  }
}

% ---------- ABSTRACT AND KEYWORDS ----------
\IEEEtitleabstractindextext{
\begin{abstract}
\justifying{
Pose transfer refers to the probabilistic image generation of a person with a previously unseen novel pose from another image of that person having a different pose. Due to potential academic and commercial applications, this problem is extensively studied in recent years. Among the various approaches to the problem, attention guided progressive generation is shown to produce state-of-the-art results in most cases. In this paper, we present an improved network architecture for pose transfer by introducing attention links at every resolution level of the encoder and decoder. By utilizing such dense multi-scale attention guided approach, we are able to achieve significant improvement over the existing methods both visually and analytically. We conclude our findings with extensive qualitative and quantitative comparisons against several existing methods on the DeepFashion dataset.
}
\end{abstract}

\begin{IEEEkeywords}
Pose transfer, Attention, GAN, DeepFashion.
\end{IEEEkeywords}
}

\maketitle

% ---------- 1. INTRODUCTION ----------
\IEEEraisesectionheading{\section{Introduction}\label{sec:introduction}}
\IEEEPARstart{S}{ynthesizing} novel views of an object or a scene can have several applications in computer graphics, image reconstruction, photography, multi-view data generation etc. While view synthesis can be useful in many practical scenarios, this is a significantly challenging problem in computer vision due to a number of factors including occlusion, illumination, geometric and perspective distortions. This is particularly difficult for deformable objects because their shapes can also vary in a specific view which leads to a much larger number of possible views to predict from a single observation.

Pose transfer can be assumed as a fine-grained view synthesis problem where we like to estimate a view (image) of a deformable object (person) with a particular state (pose) from a single observation. Therefore, pose transfer requires an exceptionally robust generative algorithm to infer both the visible and occluded body parts in a new unobserved state (target pose) from a single observed state (source pose) while preserving the general appearance of the person including expression, skin tone, dress and background.

\begin{figure}[t]
  \centering
  \includegraphics[width=\linewidth]{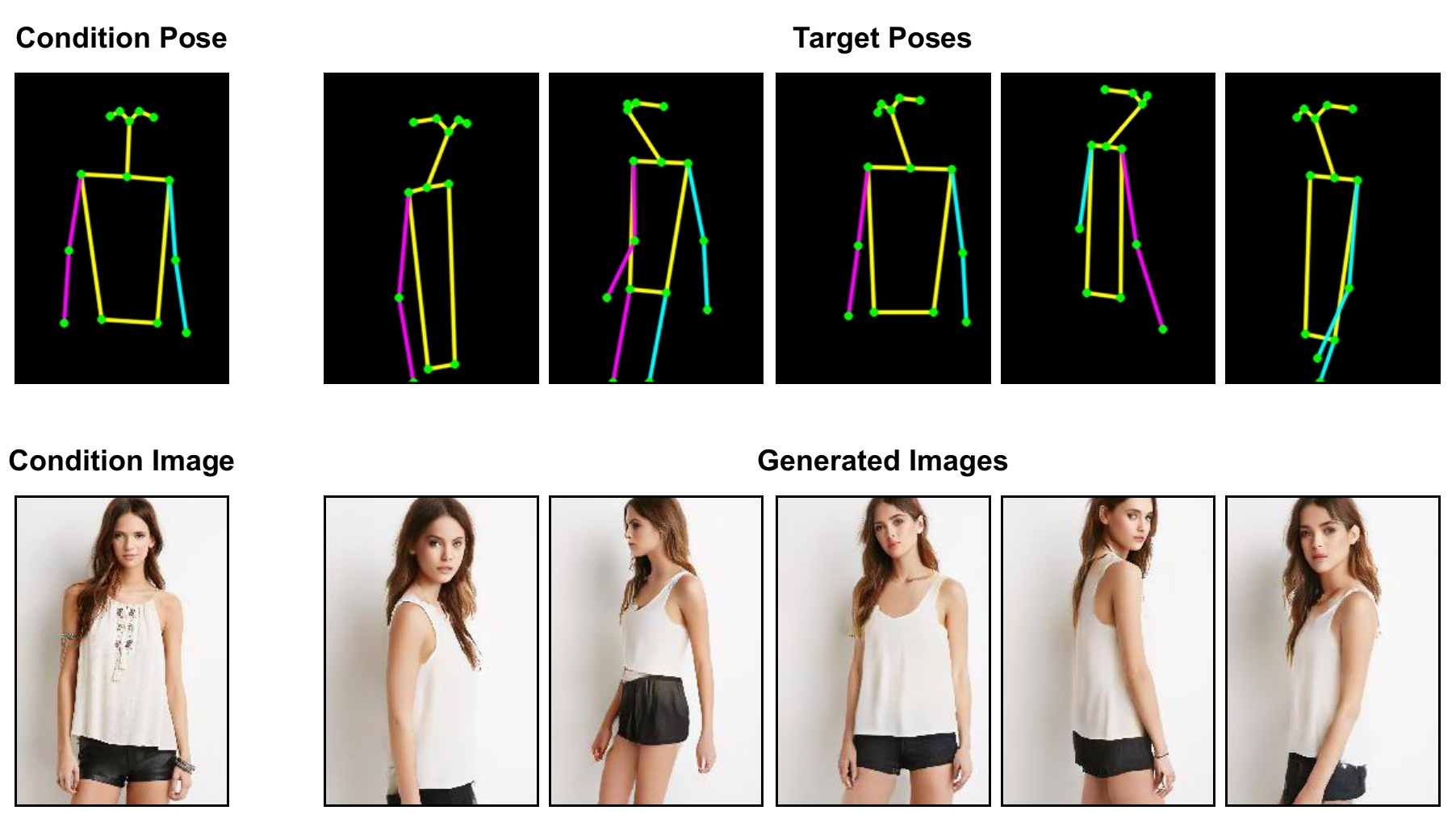}
  \caption{General overview of pose transfer using the proposed method.}
  \label{fig:introduction}
\end{figure}

In Fig. \ref{fig:introduction}, we demonstrate the general overview of pose transfer. Here, we have an image, $I_A$ of a person with an initial source pose, $P_A$. The task of the algorithm is to generate an image, $I_B$ of the same person corresponding to an unobserved target pose, $P_B$. We refer to $I_A$ and $I_B$ as \emph{condition image} and \emph{generated image} respectively.

Initial solution to this problem \cite{ma2017pose, ma2018disentangled} introduced a coarse to fine generation by dividing the problem into individual sub-tasks to handle foreground, background and pose separately. The complexity of such pipeline is later simplified to a unified approach by utilizing deformable GANs \cite{siarohin2018deformable} and variational U-Net \cite{esser2018variational}. More recently, a more streamlined approach \cite{zhu2019progressive} is introduced by leveraging attention mechanism to progressively transfer pose. The key idea is to perform pose transfer on a local manifold at each intermediate step to avoid the difficulties arise due to much complex structures on the global manifold. In this technique, the condition image and the sparse representations of the poses are initially downsampled to a lower resolution by an encoder. Then, the attention guided progressive pose transfer is performed with the encoded images. Finally, the generated image is upsampled back to higher resolution by a decoder. This method is able to achieve significantly better generation quality than all previous approaches.

This effectiveness of an attention guided approach motivates us to further explore the potential improvements of generation quality. We hypothesized that progressively transferring pose only at a lower resolution followed by upsampling cause a significant loss of finer details present in the condition image. To mitigate this information loss, we propose attention links at every underlying resolution levels of encoder and decoder. In this approach, we introduce a network architecture with such links without requiring any cascaded pose transfer block \cite{zhu2019progressive}.

\begin{figure*}[t]
  \centering
  \includegraphics[width=\linewidth]{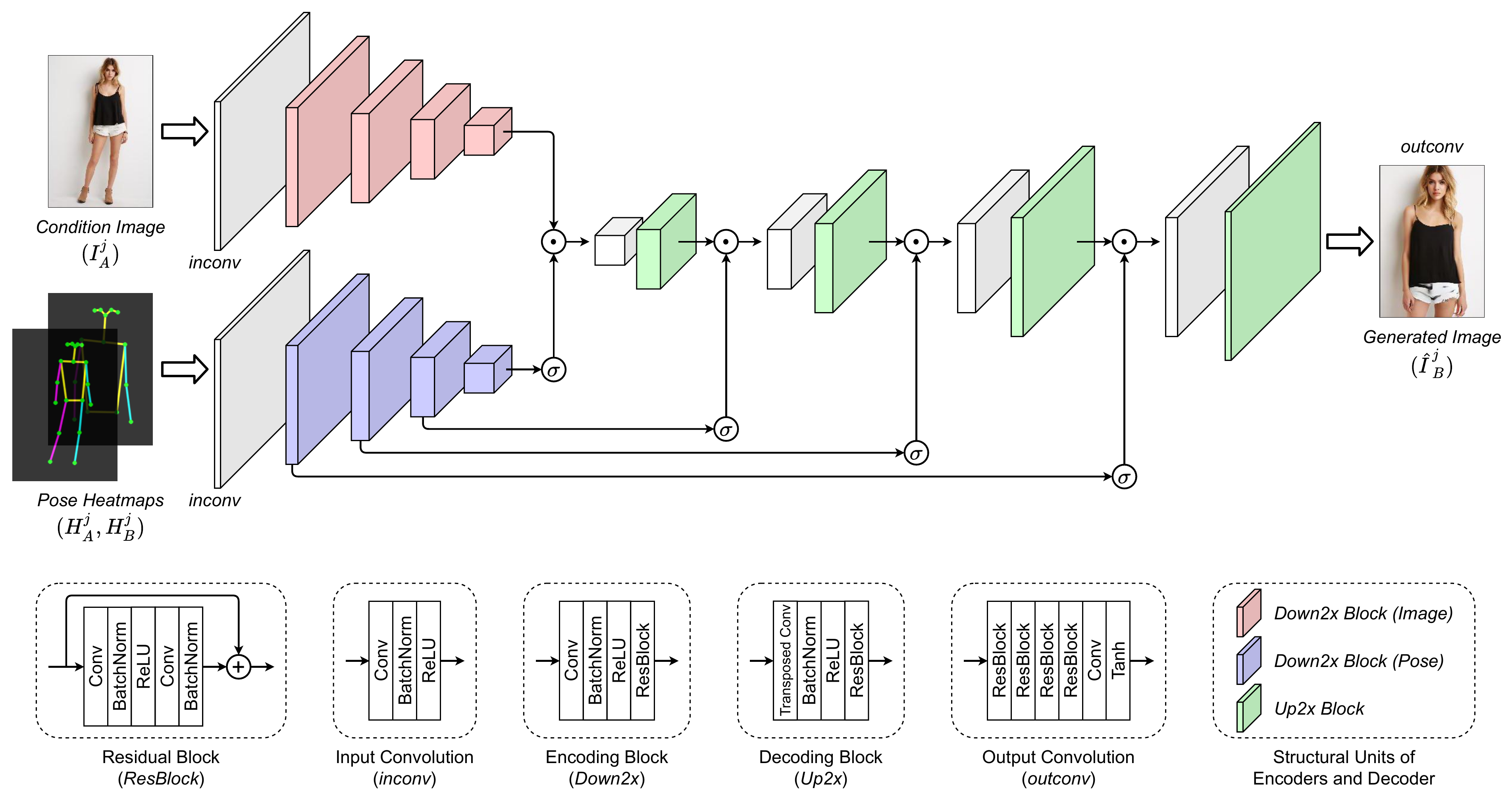}
  \caption{Architecture of the proposed generator. The generator takes the condition image $I_A^j$ along with the channel-wise concatenated pose heatmaps $(H_A^j, H_B^j)$ as inputs and generates an estimate $\hat{I}_B^j$ of the target image $I_B^j$.}
  \label{fig:generator}
\end{figure*}

In Fig. \ref{fig:generator}, we show the architecture of the proposed generator. We perform extensive comparative studies among various approaches of pose transfer on the DeepFashion dataset \cite{liu2016deepfashion}. Our method exhibits significant improvements on the generation performance in both qualitative and analytical benchmarks.

% ---------- RELATED WORKS ----------
\section{Related Works}\label{sec:relatedworks}
Image synthesis is a fundamental problem in computer vision. Since the introduction of Generative Adversarial Networks (GANs) \cite{goodfellow2014generative}, exceptional improvements in realistic image synthesis have been achieved by several variations \cite{mirza2014conditional, radford2016unsupervised, johnson2016perceptual, lassner2017generative, ledig2017photo} of the core GAN architecture in recent years. Initial proposal of GAN \cite{goodfellow2014generative} introduces unconstrained image generation by enforcing an adversarial learning scheme between a generator and a discriminator. Later, conditional GAN \cite{mirza2014conditional} extends the original idea to constrained image generation. Conditional GANs are successfully applied to several computer vision problems including image-to-image translation \cite{isola2017image, sangkloy2017scribbler, zhu2017unpaired}, inpainting \cite{yeh2017semantic}, super-resolution \cite{dong2015image, kim2016accurate}. Lassner \textit{et al.} \cite{lassner2017generative} propose a generation scheme conditioned over a variational auto-encoder (VAE) to render a 3D model of a person as a fully clothed image. Zhao \textit{et al.} \cite{zhao2018multi} propose a method for generating multi-view images of a person from a single observation in a coarse to fine approach. Balakrishnan \textit{et al.} \cite{balakrishnan2018synthesizing} introduce a pose transfer scheme to segment and generate foreground and background separately. A similar computer vision problem of \textit{Virtual Try On} is also introduced where a human model can be conditionally rendered with a given set of apparel. Han \textit{et al.} \cite{han2018viton} propose a thin plate spline transformation guided generation to transfer a specific cloth onto a person in an image. Wang \textit{et al.} \cite{wang2018toward} propose a geometric matching module in a characteristic preserving generative network.

Initial approach to pose transfer \cite{ma2017pose} introduces a two stage method in a coarse to fine manner. This technique is further improved in a disentangled person image generation scheme \cite{ma2018disentangled} by introducing separate branches for foreground, background and pose in the network architecture. In \cite{zanfir2018human}, authors propose a pose transfer method by estimating a 3D mesh from a single image. In \cite{pumarola2018unsupervised}, authors propose an unsupervised multi-level generation method with a pose conditioned bidirectional generator. Siarohin \textit{et al.} \cite{siarohin2018deformable} propose deformable GANs with nearest neighbour loss for pose transfer. Esser \textit{et al.} \cite{esser2018variational} propose a variational U-Net architecture to generate different poses of an object. More recently, an attention guided progressive generation scheme \cite{zhu2019progressive} is introduced by leveraging attention mechanism to progressively transfer pose. In this approach, pose transfer is performed on a local manifold at each intermediate step to overcome several difficulties arise due to much complex structures on the global manifold.

% ---------- PROPOSED METHOD ----------
\section{Proposed Method}\label{sec:proposedmethod}
Let us assume that we have a condition image $I_A^j$ of a person with pose $P_A^j$, where $j$ denotes the index of the person in the dataset. Our goal is to generate an image $I_B^j$ of the same person with a new pose $P_B^j$. Similar to the previous approaches \cite{ma2017pose, ma2018disentangled, siarohin2018deformable, zhu2019progressive}, we represent a pose as a set of 2D coordinates of body keypoints. We employ a pre-trained Human Pose Estimator (HPE) \cite{cao2017realtime} to estimate 18 keypoints on a human body to represent each pose. The HPE estimates each keypoint as a triplet $(x_i, y_i, v_i)$ where, $(x_i, y_i)$ represents the 2D coordinate and $v_i$ denotes a binary state for visibility of the $i$-th keypoint. More specifically, the value of $v_i$ is 1 for visible keypoints and 0 for occluded keypoints. To represent a pose as a sparse heatmap, we construct a $(h \times w \times 18)$ tensor where, $h$ and $w$ denotes the height and width of the corresponding person image respectively. Each of the 18 channels of the tensor corresponds to one specific keypoint. For a visible keypoint $(x_k, y_k, 1)$, we put a value of 1 at the location corresponds to the spatial coordinate $(x_k, y_k)$ on the $k$-th channel of the tensor. We put a value of 0 everywhere else in the tensor. We denote the sparse pose heatmaps as $H_A^j$ and $H_B^j$ corresponding to the poses $P_A^j$ and $P_B^j$ respectively.

The proposed generator takes the RGB condition image $I_A^j$ of dimension $(h \times w \times 3)$ and the channel-wise concatenated pose heatmaps $H_A^j$ and $H_B^j$ of dimension $(h \times w \times 36)$ as inputs and generates an RGB image $\hat{I}_B^j$. We employ a PatchGAN discriminator \cite{isola2017image} for determining visual correctness of the generated images. The discriminator takes two channel-wise concatenated RGB images, either $(I_A^j, I_B^j)$ or $(I_A^j, \hat{I}_B^j)$, of dimension $(h \times w \times 6)$ as input and estimates a binary class probability map for the input patches.

\subsection{Generator}\label{sec:generator}
In Fig. \ref{fig:generator}, we illustrate the architecture of the proposed generator. The generator is composed of two downsampling paths (encoders) followed by an upsampling path (decoder) with attention links between feature maps at every underlying resolution levels. We begin by describing the essential components of the generator followed by encoders, decoder and the attention mechanism.

\subsubsection{Generator Components}\label{sec:generatorcomponents}
\textbf{Conv1x1.} A point-wise 2D convolution operation that preserves the size of the input. We perform a single 2D convolution with $1 \times 1$ kernel, stride = 1, padding = 0 and without adding any bias.\\\\
\textbf{Conv3x3.} A 2D convolution operation that preserves the size of the input. We perform a single 2D convolution with $3 \times 3$ kernel, stride = 1, padding = 1 and without adding any bias.\\\\
\textbf{Residual Block.} A basic residual block \cite{he2016deep} that preserves the size of the input along with the number of channels. A residual block is composed of 5 sequential layers -- Conv3x3, Batch Normalization \cite{ioffe2015batch}, ReLU \cite{nair2010rectified}, Conv3x3, Batch Normalization. The input is passed through these layers and the result is added with the original input to produce the final output.\\\\
\textbf{Down2x Block.} A downsampling block in the encoder for compressing the input size by a factor of 2. We perform a 2D convolution with $4 \times 4$ kernel, stride = 2, padding = 1 and without adding any bias to downsample the input. The resulting feature maps are passed through sequential layers of Batch Normalization, ReLU and a Residual Block to produce the final output.\\\\
\textbf{Up2x Block.} An upsampling block in the decoder for expanding the input size by a factor of 2. We perform a 2D transposed convolution with $4 \times 4$ kernel, stride = 2, padding = 1 and without adding any bias to upsample the input. The resulting feature maps are passed through sequential layers of Batch Normalization, ReLU and a Residual Block to produce the final output.

\subsubsection{Encoders}\label{sec:encoders}
In our approach, we have two parallel downstream branches in the encoder. The \emph{image branch} corresponds to the condition image $I_A^j$ and the \emph{pose branch} corresponds to the concatenated pose heatmaps $(H_A^j$, $H_B^j)$. Initially, we perform a 2D convolution operation Conv3x3 to project each input into a feature space of $(h \times w \times N_f)$ where, $N_f$ denotes the initial number of feature maps. The convolution is followed by Batch Normalization and ReLU activation to produce the initial input feature maps at each branch. The input feature maps are then passed through $N$ subsequent Down2x Blocks at each branch. At each Down2x Block, we downsampled the input size by a factor of 2 while expanding the number of feature maps by a factor of 2. Therefore, after $N$ consecutive downsampling blocks, we ended up with a feature space of $(\frac{h}{2^N} \times \frac{w}{2^N} \times N_f * 2^N)$.

\subsubsection{Decoder}\label{sec:decoder}
In our approach, we have a single upstream branch in the decoder for generating the image $\hat{I}_B^j$ with the target pose. Starting with a feature space of dimension $(\frac{h}{2^N} \times \frac{w}{2^N} \times N_f * 2^N)$, we pass the feature maps through $N$ consecutive Up2x Blocks. At each Up2x Block, we upsampled the input size by a factor of 2 while compressing the number of feature maps by a factor of 2. Therefore, after $N$ subsequent upsampling blocks, we ended up with a feature space of $(h \times w \times N_f)$. Finally, the feature maps are passed through 4 consecutive Residual Blocks followed by a point-wise 2D convolution operation Conv1x1 to project the feature space into an output of dimension $(h \times w \times 3)$. We apply the hyperbolic tangent activation function $tanh$ on the output tensor to get the normalized generated image $\hat{I}_B^j$.

\subsubsection{Attention Mechanism}\label{sec:attentionmechanism}
We construct attention links between downstream and upstream branches at each resolution level. For resolution level $k$, we compute the attention mask $M_k$ by applying an element-wise $sigmoid$ activation function $\sigma$ on the encoded feature maps of the \emph{pose branch} $H_k^\mathcal{E}$. The image feature maps $I_k$ at resolution level $k$ are updated by performing an element-wise product with the attention mask $M_k$. The updated image feature maps $I_k$ act as the input to the upsampling block of the decoder at resolution level $k$. We repeat these operations sequentially up to the highest resolution which results $N$ such attention links. Mathematically, at the lowest resolution level where, $k = N$,
\begin{equation*}
  I_{N-1}^\mathcal{D} = \mathcal{D}_N^{Up2x} (I_N^\mathcal{E} \;\odot\; \sigma(H_N^\mathcal{E}))
\end{equation*}
and for each subsequent higher resolution level where, $k = \{1, ..., N-1\}$,
\begin{equation*}
  I_{k-1}^\mathcal{D} = \mathcal{D}_k^{Up2x} (I_k^\mathcal{D} \;\odot\; \sigma(H_k^\mathcal{E}))
\end{equation*}
Here, we denote downstream encoding as $\mathcal{E}$ and upstream decoding as $\mathcal{D}$.

\subsection{Discriminator}\label{sec:discriminator}
We use a Markovian PatchGAN discriminator \cite{isola2017image} to estimate high-frequency correctness in the generated images. The discriminator complements to the L1 loss which is restricted to low-frequency details only. Such a discriminator operates on $S \times S$ image patches by classifying each patch as real or fake. As explained by the authors \cite{isola2017image}, PatchGAN functions as a style/texture loss by modeling the image as a Markov random field, assuming independence between pixels separated by more than a patch diameter.

\begin{figure}[t]
  \centering
  \includegraphics[width=\linewidth]{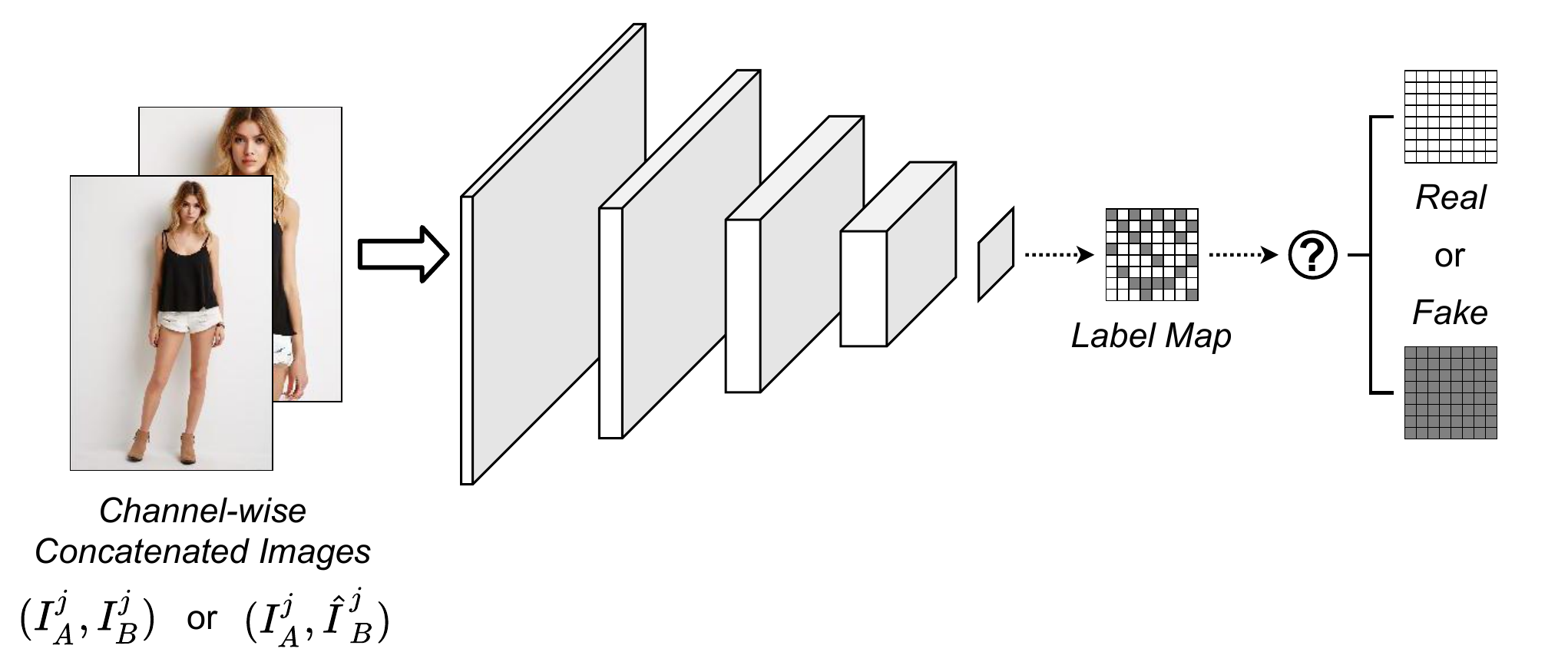}
  \caption{Architecture of the PatchGAN discriminator. The discriminator takes two channel-wise concatenated images, either $(I_A^j, I_B^j)$ or $(I_A^j, \hat{I}_B^j)$, as input and estimates a label map where, each label corresponds to the binary class probability of an input patch.}
  \label{fig:disciminator}
\end{figure}

In our approach, we enforce adversarial discrimination on the image transition rather than the image itself. We do this by depth-wise concatenating the condition image $I_A^j$ either with the target image $I_B^j$ or with the generated image $\hat{I}_B^j$ where, $(I_A^j, I_B^j)$ is labeled as real and $(I_A^j, \hat{I}_B^j)$ is labeled as fake. We adopt an identical network architecture as \cite{isola2017image} that effectively operates on a $70 \times 70$ receptive field (patch) of the input. In Fig. \ref{fig:disciminator}, we illustrate the architecture of the PatchGAN discriminator.

\subsection{Training}\label{sec:training}
We train our network in an adversarial scheme where, we have two competing objective functions for the generator $G$ and the discriminator $D$. During training, each objective tries to minimize the penalty incurred by itself while trying to maximize the penalty for the other one.

\subsubsection{Generator Objective}\label{sec:generatorobjective}
To ensure the low-frequency correctness in the generated images, we compute $L_1$ loss between the target image $I_B^j$ and the generated image $\hat{I}_B^j$. Mathematically,

\begin{equation*}
  \mathcal{L}_1^G = \|\hat{I}_B^j - I_B^j\|_1
\end{equation*}

To estimate the high-frequency correctness in the generated images, we compute Binary Cross-Entropy (BCE) loss using the PatchGAN discriminator. Mathematically,

\begin{equation*}
  \mathcal{L}_{GAN}^G = \mathcal{L}_{BCE}(D(I_A^j, \;\hat{I}_B^j), \;1)
\end{equation*}

To improve the visual fidelity of the generated images, we also include a perceptual loss \cite{johnson2016perceptual} in our generator objective. Mathematically,

\begin{equation*}
  \mathcal{L}_{P_\rho}^G = \frac{1}{h_\rho w_\rho c_\rho} \sum_{x=1}^{h_\rho} \sum_{y=1}^{w_\rho} \sum_{z=1}^{c_\rho} \|\phi_\rho(\hat{I}_B^j) - \phi_\rho(I_B^j)\|_1
\end{equation*}
where, $\mathcal{L}_{P_\rho}^G$ denotes the perceptual loss computed from the output of the $\rho^{\text{th}}$ layer of a pre-trained VGG19 model \cite{simonyan2015very}, $\phi_\rho$ denotes the output of the $\rho^{\text{th}}$ layer having a feature space of dimension $(h_\rho \times w_\rho \times c_\rho)$. In our approach, we compute the perceptual loss at two different layers ($4^{\text{th}}$ and $9^{\text{th}}$) of a VGG19 model pre-trained on the ImageNet dataset \cite{deng2009imagenet} to capture both coarse and fine details in the generated images.

The complete generator objective is calculated as a weighted linear combination of $L_1$ loss, GAN loss and perceptual loss. Mathematically,

\begin{equation*}
  \mathcal{L}^G = \text{arg} \min_{G} \max_{D} \;\;\lambda_1 \mathcal{L}_1^G + \lambda_2 \mathcal{L}_{GAN}^G + \lambda_3 (\mathcal{L}_{P_4}^G + \mathcal{L}_{P_9}^G)
\end{equation*}
where, $\lambda_1$, $\lambda_2$ and $\lambda_3$ denote the weights for the corresponding loss functions.

\subsubsection{Discriminator Objective}\label{sec:discriminatorobjective}
Our discriminator objective has a single GAN loss component which is calculated as the average BCE loss over a real image transition $(I_A^j, I_B^j)$ and a fake image transition $(I_A^j, \hat{I}_B^j)$. Mathematically,

\begin{equation*}
  \mathcal{L}_{GAN}^D = \frac{1}{2} \left[\mathcal{L}_{BCE}(D(I_A^j, \;I_B^j), \;1) + \mathcal{L}_{BCE}(D(I_A^j, \;\hat{I}_B^j), \;0)\right]
\end{equation*}
where, we assume real label as 1 and fake label as 0. The complete discriminator objective is given by,

\begin{equation*}
  \mathcal{L}^D = \text{arg} \min_{D} \max_{G} \;\;\mathcal{L}_{GAN}^D
\end{equation*}

\subsection{Implementation}\label{sec:implementation}
In our implementation, the generator is constructed with 4 downsampling blocks in both the encoders and consequently 4 upsampling blocks in the decoder $(N = 4)$. The initial number of feature maps is set to 64 $(N_f = 64)$. In the generator objective, we set the weights as, $\lambda_1 = 5$, $\lambda_2 = 1$ and $\lambda_3 = 5$. We initialize the parameters of both generator and discriminator before training by sampling from a normal distribution of 0 mean and 0.02 standard deviation. We optimize both generator and discriminator using the stochastic Adam optimizer \cite{kingma2015adam} with learning rate $\eta = 1e^{-3}$, $\beta_1 = 0.5$, $\beta_2 = 0.999$, $\epsilon = 1e^{-8}$ and weight decay = 0. We train the network on a single NVIDIA TITAN X GPU for 270K iterations with a batch size of 8 and serialize the network weights after every 500 iterations. During inference, we select the checkpoint with best evaluation metrics among 70 most recent checkpoints.

% ---------- EXPERIMENTS ----------
\section{Experiments}\label{sec:experiments}
To evaluate the performance of the proposed architecture, we carry out extensive qualitative and quantitative comparisons against a number of previous major pose transfer methods \cite{ma2017pose, siarohin2018deformable, esser2018variational, zhu2019progressive} on a fixed dataset \cite{liu2016deepfashion}. In most of the evaluation metrics, our method outperforms the previous methods. We also perform user studies for a subjective visual quality assessment.\\\\
\textbf{Dataset:} In our experiments, we use DeepFashion \cite{liu2016deepfashion} \textit{In-shop Clothes Retrieval} dataset for performance evaluation and comparison with previous methods. The dataset contains $176 \times 256$ person images centered on $256 \times 256$ square grids. It features a wide variation of outfit and pose which makes this dataset a relevant choice for evaluating and comparing pose transfer algorithms. For direct and fair comparison, we adopt the exact train-test split provided by Zhu \textit{et al.} \cite{zhu2019progressive} where, 101,966 image pairs are randomly selected for training and 8,570 image pairs for testing. Also for a better generalization, the image pairs are selected such that the person identities of the training set do not overlap with those of the testing set.\\\\
\textbf{Metrics:} At present, a quantifiable generalized metric for visual quality assessment of images is an open problem in computer vision. However, in the previous pose transfer algorithms \cite{ma2017pose, siarohin2018deformable, esser2018variational, zhu2019progressive}, authors estimate a few widely used evaluation metrics for quantifying visual quality. This includes Structural Similarity Index (SSIM) \cite{wang2004image}, Inception Score (IS) \cite{salimans2016improved}, Detection Score (DS) \cite{liu2016ssd} and PCKh \cite{andriluka20142d}. SSIM measures the perceived quality of generated images by comparing with respective real images and considering image degradation as perceived change in structural information. IS uses Inception architecture \cite{szegedy2015going} as an image classifier to estimate the KL divergence \cite{kullback1951information} between the label distribution and the marginal distribution for a large set of images. DS uses an object detector to estimate the target class recognition confidence of the object detection model as a measure of the perceptual quality. PCKh aims to quantify the shape consistency between generated and real person images by estimating the percentage of correctly aligned keypoints. We also evaluate the Learned Perceptual Image Patch Similarity (LPIPS) \cite{zhang2018unreasonable} metric which is a more modern standard for perceptual image quality assessment.

\subsection{Qualitative and Quantitative Comparison}\label{sec:comparison}
\begin{figure}[t]
  \centering
  \includegraphics[width=\linewidth]{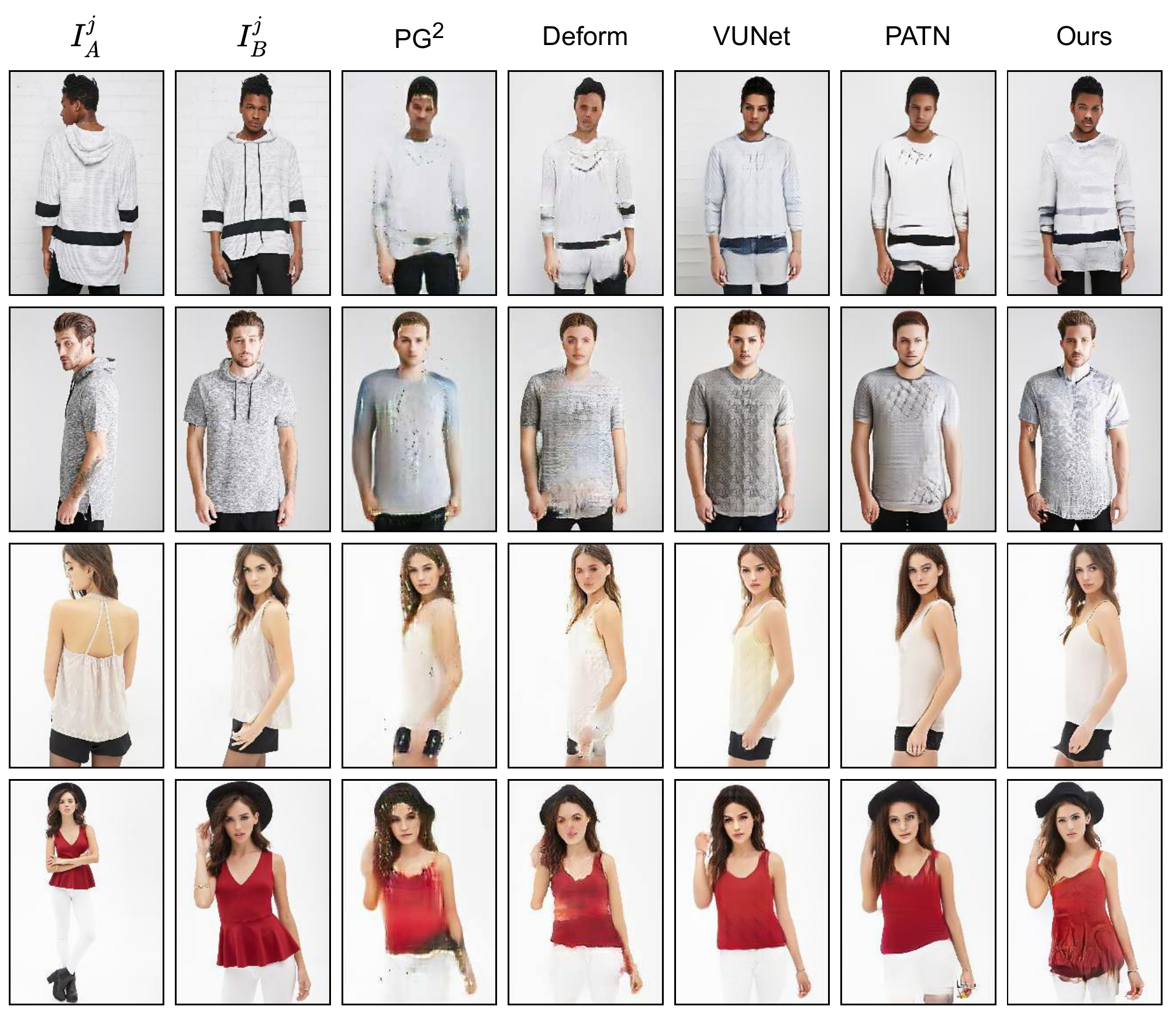}
  \caption{Qualitative comparison among different pose transfer methods. $I_A^j$ denotes the condition image, $I_B^j$ denotes the target image and subsequent columns show the generated images by $\text{PG}^2$ \cite{ma2017pose}, Deformable GANs \cite{siarohin2018deformable}, VUNet \cite{esser2018variational}, PATN \cite{zhu2019progressive} and our method.}
  \label{fig:comparison}
\end{figure}

In Fig. \ref{fig:comparison}, we show a qualitative comparison among previously proposed major pose transfer algorithms \cite{ma2017pose, siarohin2018deformable, esser2018variational, zhu2019progressive} and our method. For a direct and fair comparison we use the same image pairs $(I_A^j, I_B^j)$ as \cite{zhu2019progressive} and expanded their visual analysis by incorporating results of our method. It can be observed that images generated by our method are able to retain better skin color, hair style, facial hair and limb structure. Our method also preserves dress texture better than other methods which is more prominent in 2nd and 4th row in Fig. \ref{fig:comparison}. From a generic visual inspection, generated images by our method apparently look more realistic compared to other methods.

The apparent visual superiority of our method is further reflected in quantitative evaluation of multiple perceptual metrics. In Table \ref{tab:comparison}, we show an analytical comparison among different pose transfer methods by evaluating SSIM, IS, DS, PCKh and LPIPS scores. To ensure cycle consistency, we generate images in both directions -- $\hat{I}_B^j$ from the image pairs $(I_A^j, I_B^j)$ followed by $\hat{I}_A^j$ from the reverse image pairs $(I_B^j, I_A^j)$. We evaluate each metric on both $\hat{I}_B^j$ and $\hat{I}_A^j$ followed by estimating their mean as the final score.
For SSIM and IS, we obtain slightly lower scores than the best results. For DS, PCKh and LPIPS, our method is able to achieve best scores. We improve the PCKh score over PATN \cite{zhu2019progressive} by 2\% indicating a superior shape consistency with better alignment among keypoints. Similar to \cite{gafni2020wish}, we evaluate LPIPS score for both VGG16 \cite{simonyan2015very} and SqueezeNet \cite{iandola2016squeezenet}. In each case, our method achieves significantly better score indicating improved perceptual quality over other methods.

\begin{table}[t]
\centering
\caption{Quantitative comparison among different pose transfer methods.}
\begin{tabular}{l|cccccc}
\hline
Method & SSIM & IS & DS & PCKh & \begin{tabular}[c]{@{}c@{}}LPIPS\\ (VGG)\end{tabular} & \begin{tabular}[c]{@{}c@{}}LPIPS\\ (SqzNet)\end{tabular} \\ \hline
$\text{PG}^2$ \cite{ma2017pose}      & 0.773 & 3.163 & 0.951 & 0.89 & 0.523 & 0.416 \\
Deform \cite{siarohin2018deformable} & 0.760 & 3.362 & 0.967 & 0.94 & -     & -     \\
VUNet \cite{esser2018variational}    & 0.763 & \textbf{3.440} & 0.972 & 0.93 & -     & -     \\
PATN \cite{zhu2019progressive}       & \textbf{0.773} & 3.209 & 0.976 & 0.96 & 0.299 & 0.170 \\
Ours                                 & 0.769 & 3.379 & \textbf{0.976} & \textbf{0.98} & \textbf{0.200} & \textbf{0.111} \\ \hline
Real Data                            & 1.000 & 3.864 & 0.974 & 1.00 & 0.000 & 0.000 \\ \hline
\end{tabular}
% \caption{Quantitative comparison among different pose transfer methods.}
\label{tab:comparison}
\end{table}

\subsection{User Study}\label{sec:userstudy}
Although the evaluation metrics discussed so far are widely used for quantifying visual quality, a sufficiently large number of outliers show the limitations of these metrics to be able to generalize the subjective nature of such assessment. Therefore, human perception is perhaps the most reliable way to assess visual quality. For this reason, we perform an opinion based user study to analyze the visual quality of the generated images as perceived by humans. We include two tracks in this study -- a constrained test followed by an unconstrained test. In the constrained test, users need to perform discrimination between real and fake images within a fixed amount of time. In the unconstrained test, users need to perform similar discrimination but without any time limit. We follow a similar protocol as \cite{ma2017pose, siarohin2018deformable, zhu2019progressive} for the constrained test except the allowed exposure time for each image. We increase the exposure time from 1 second to 5 second which allows users a better observation before making a decision. This also puts our method in a more challenging position compared to \cite{ma2017pose, siarohin2018deformable, zhu2019progressive}.

For our user study, we select 130 real and 130 generated images. 10 images from each set are used as the fixed practice samples. A user is shown 20 images during a test where, 10 images are randomly drawn from each set of the remaining 120 images. For every anonymous user submission, the fraction of real images identified as generated (\textbf{R2G}) and the fraction of generated images identified as real (\textbf{G2R}) are recorded. The final score is calculated as the mean of global aggregation of all submissions. In Table \ref{tab:userstudy}, we show the evaluation scores of our user study conducted with 61 individuals along with the scores reported in previous works \cite{ma2017pose, siarohin2018deformable, zhu2019progressive}. Even with higher exposure time for better visual observation, our method exhibits significantly higher R2G and G2R scores and consequently, much lower recognition accuracy by the users. These results further imply that the images generated by our method are visually more realistic than other methods leading to higher confusion among the users during discrimination.

\begin{table}[t]
\centering
\caption{Evaluation scores of user study for different pose transfer methods.}
\begin{tabular}{l|cccc}
\hline
Method &
  \begin{tabular}[c]{@{}c@{}}Exposure Time\\ (second)\end{tabular} &
  \begin{tabular}[c]{@{}c@{}}R2G\\ (\%)\end{tabular} &
  \begin{tabular}[c]{@{}c@{}}G2R\\ (\%)\end{tabular} &
  \begin{tabular}[c]{@{}c@{}}Accuracy\\ (\%)\end{tabular} \\ \hline
$\text{PG}^2$ \cite{ma2017pose}      & 1.0 & ~9.20 & 14.90 & 87.95 \\
Deform \cite{siarohin2018deformable} & 1.0 & 12.42 & 24.61 & 81.49 \\
PATN \cite{zhu2019progressive}       & 1.0 & 19.14 & 31.78 & 74.54 \\
Ours                                 & 5.0 & 25.90 & \textbf{54.26} & \textbf{59.92} \\
Ours                                 & $\infty$ & \textbf{30.37} & 46.30 & 61.67 \\ \hline
\end{tabular}
% \caption{Evaluation scores of user study for different pose transfer methods.}
\label{tab:userstudy}
\end{table}

\subsection{Ablation Study}\label{sec:ablationstudy}
The core design characteristic of the proposed network architecture is focused around the attention links at different underlying resolution levels of the generator as shown in Fig. \ref{fig:generator}. To study the efficacy of such architecture, we perform an ablation study with 4 variants of the generator. In the first variant, denoted as \textbf{A0}, we remove every attention link from the generator and thereby making it a generic encoder-decoder network. In the second variant, denoted as \textbf{A1-LR}, we keep only one attention link at the lowest resolution level. For the third variant, denoted as \textbf{A1-HR}, we keep only one attention link at the highest resolution level. For the fourth variant, denoted as \textbf{FULL}, we retain every attention link present in the generator as shown in Fig. \ref{fig:generator}. We train each network on the same training data for 125K iterations keeping the discriminator, training mechanism and all other implementation conditions same as discussed in Sec. \ref{sec:proposedmethod}.

\begin{figure}[t]
  \centering
  \includegraphics[width=\linewidth]{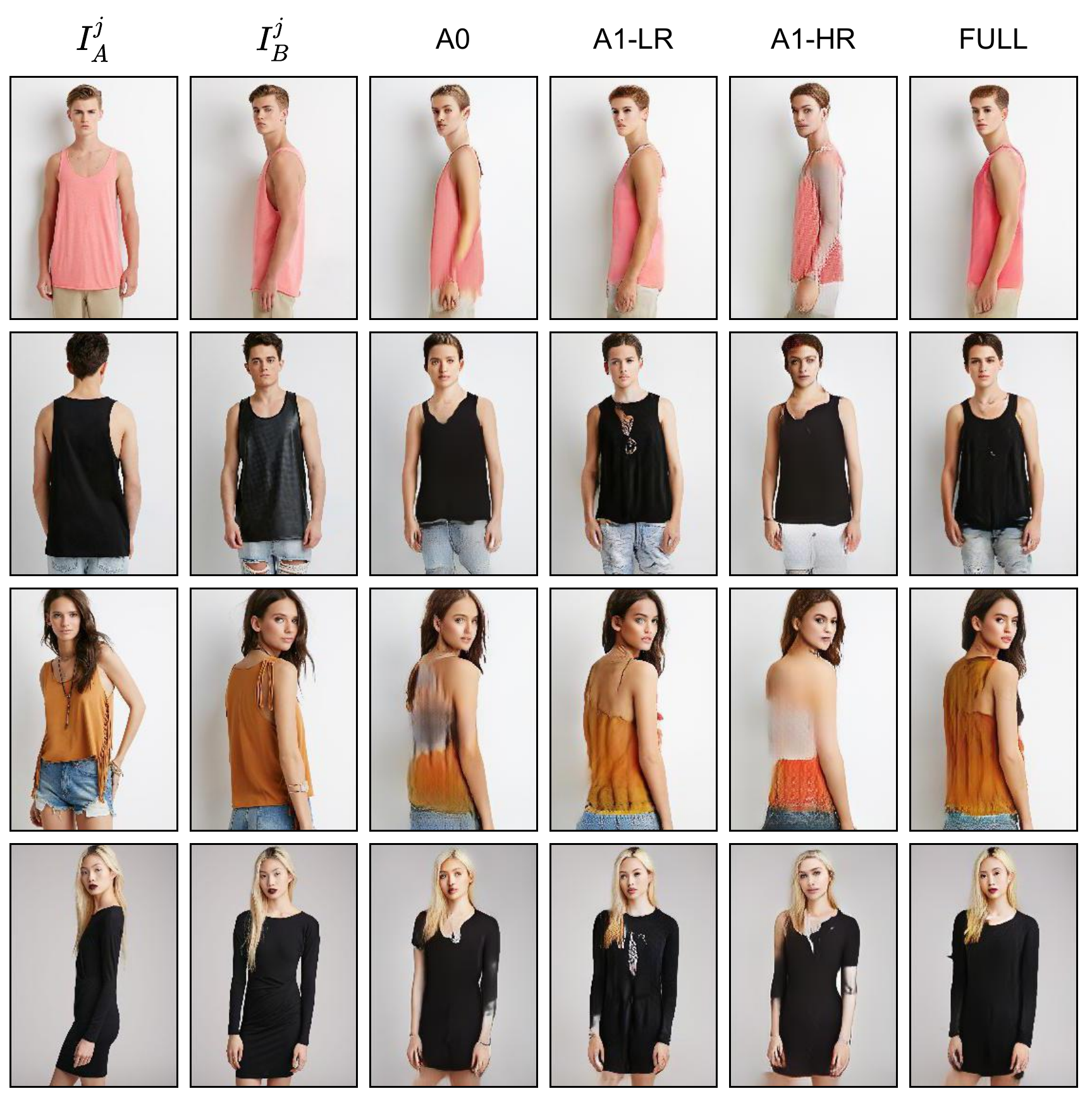}
  \caption{Ablation study -- Qualitative comparison among model variants. $I_A^j$ denotes the condition image, $I_B^j$ denotes the target image and each subsequent column shows the generated images by respective model variant.}
  \label{fig:ablation}
\end{figure}

In Fig. \ref{fig:ablation}, we show a qualitative comparison among the generated images by different model variants. Model A0 generates blurry and visually inconsistent images. Model A1-LR generates visually and structurally consistent images but lacks realistic detail in the face and limbs. Model A1-HR performs very poorly resulting in blurry and incomplete images. Model FULL performs significantly better than other variants producing realistic images with finer detail while preserving visual and structural consistency.

The subjective visual analysis is further supported by a quantitative comparison among the model variants as shown in Table \ref{tab:ablation} where, we evaluate SSIM, IS, DS, PCKh and LPIPS scores for each model. For most of the metrics, model FULL achieves best evaluation scores as expected. Interestingly, model A1-LR achieves marginally better IS and DS scores than model FULL even with visually inferior generation results as evident from Fig. \ref{fig:ablation}. This anomaly further indicates the limitations of present perceptual metrics to be able to properly generalize human perception. For this reason, the outcome of such analysis is based on multiple metrics rather than relying on a single one.

From the visual and analytical ablation study, we conclude that the dense attention links between every underlying resolution levels of encoder and decoder in the generator network result in significant improvements of visual quality of the generated images. Consequently, this leads to the architecture of the proposed generator as shown in Fig. \ref{fig:generator}.

\begin{table}[t]
\centering
\caption{Ablation study -- Quantitative comparison among model variants.}
\begin{tabular}{l|cccccc}
\hline
Model & SSIM & IS & DS & PCKh & \begin{tabular}[c]{@{}c@{}}LPIPS\\ (VGG)\end{tabular} & \begin{tabular}[c]{@{}c@{}}LPIPS\\ (SqzNet)\end{tabular} \\ \hline
A0        & 0.760 & 3.055 & 0.969 & 0.97 & 0.221 & 0.124 \\
A1-LR     & 0.758 & \textbf{3.211} & \textbf{0.976} & 0.97 & 0.210 & 0.117 \\
A1-HR     & 0.755 & 2.859 & 0.965 & 0.95 & 0.229 & 0.134 \\
FULL      & \textbf{0.764} & 3.171 & 0.975 & \textbf{0.97} & \textbf{0.204} & \textbf{0.113} \\ \hline
Real Data & 1.000 & 3.864 & 0.974 & 1.00 & 0.000 & 0.000 \\ \hline
\end{tabular}
% \caption{Ablation study -- Quantitative comparison among model variants.}
\label{tab:ablation}
\end{table}

\subsection{Failure Cases}\label{sec:failurecases}
\begin{figure}[t]
  \centering
  \includegraphics[width=\linewidth]{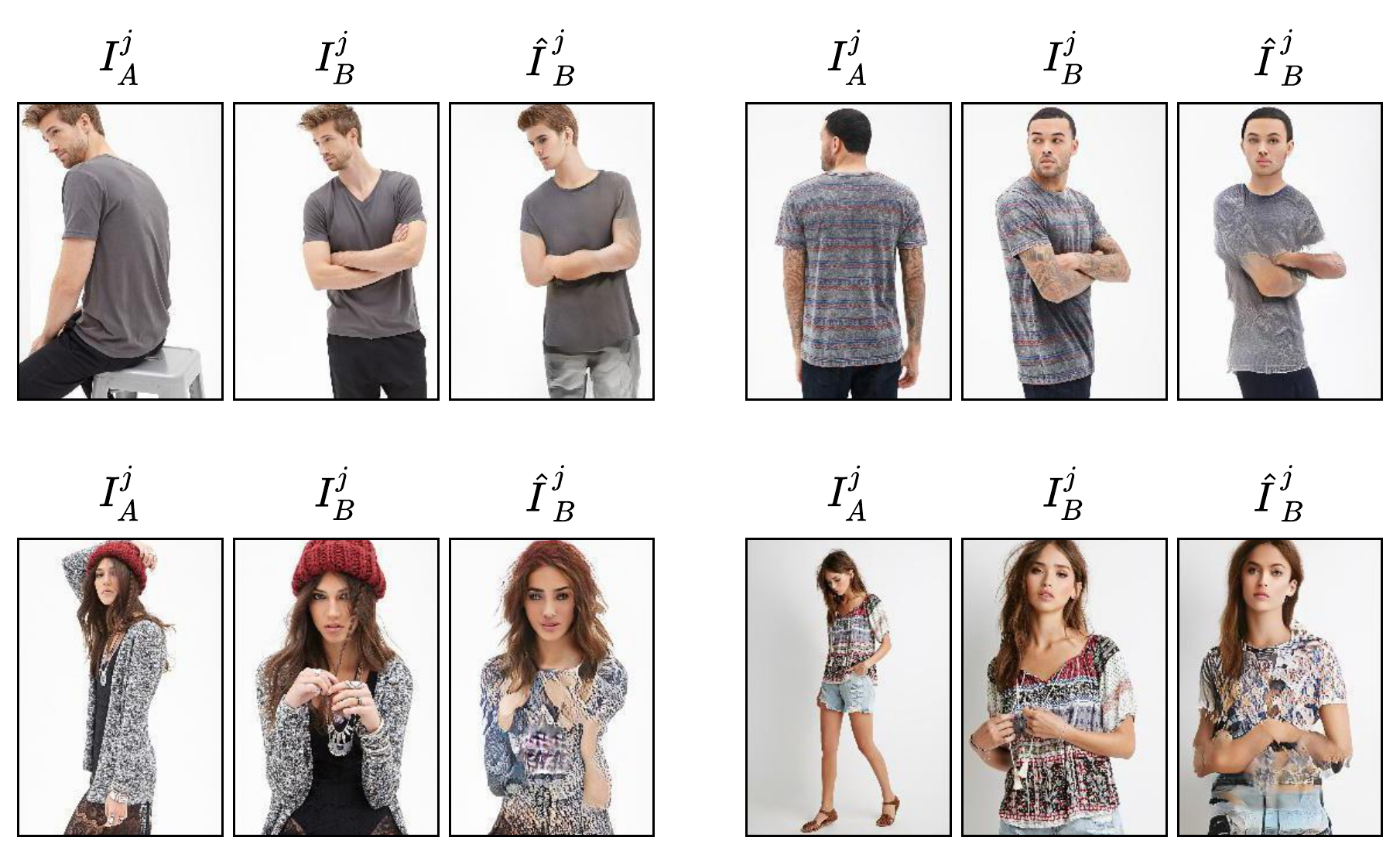}
  \caption{Failure cases. For each example, $I_A^j$ denotes the condition image, $I_B^j$ denotes the target image and $\hat{I}_B^j$ denotes the generated image.}
  \label{fig:failure}
\end{figure}

Similar to the previous keypoint based pose transfer methods \cite{ma2017pose, ma2018disentangled, siarohin2018deformable, zhu2019progressive}, performance of the proposed method is directly dependent on the success of an HPE such as \cite{cao2017realtime} in estimating the keypoints correctly. An erroneously estimated pose directly impacts the generation performance leading to inconsistency in the generated images. The supervised learning approach during optimization is also associated with the usual problem of data imbalance. In some occasions, the visual quality of the generated image is inconsistent for a distinctly uncommon pose or outfit with very few image samples in the training dataset. In Fig. \ref{fig:failure}, we show some examples where, the proposed method fails to maintain consistency in the generated images.

\subsection{Extended Applications}\label{sec:applications}
To show the efficacy of the proposed network architecture across multiple application domains, we extend our experiments to address a few different problems. An important point to note that we do not modify the network architecture in these experiments. While domain specific modifications of the architecture may further improve generation quality, those analyses are beyond the scope of this paper. However, we show that even without any modification, the proposed network works remarkably well in many different practical scenarios.

\subsubsection{Conditional Semantic Reconstruction}\label{sec:application1}

\begin{figure}[b]
  \centering
  \includegraphics[width=\linewidth]{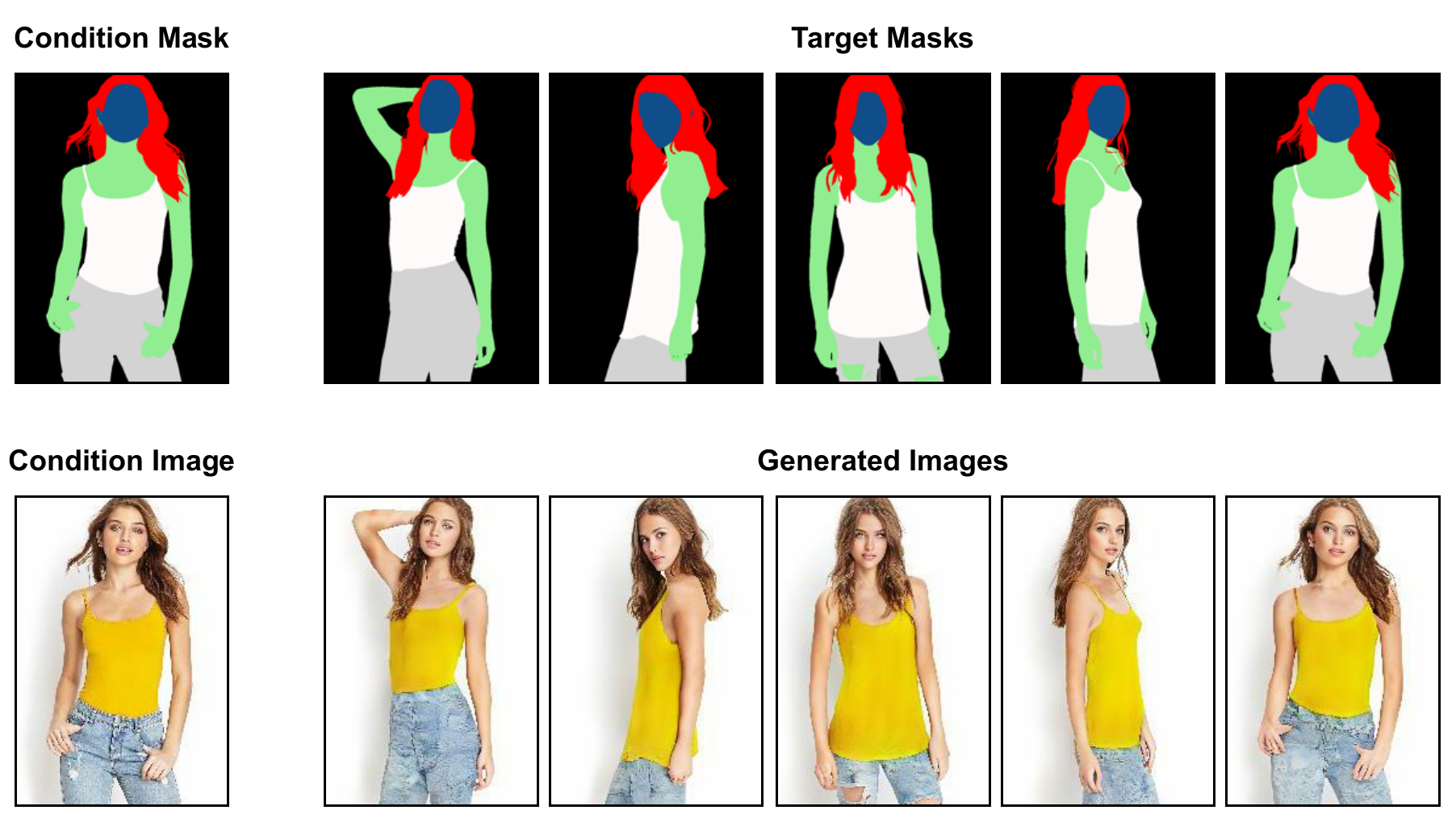}
  \caption{Qualitative results of conditional semantic reconstruction using the proposed method.}
  \label{fig:semantic}
\end{figure}

In novel view synthesis, one typical solution is image reconstruction from the crude semantic map. Unlike an unconditional synthesis, here we use a condition image as reference to transfer various image attributes to the generated image. In our experiments, we use semantic maps provided with the DeepFashion \cite{liu2016deepfashion} \textit{In-shop Clothes Retrieval} dataset. Here, a semantic map is represented as an aggregation of 16 different parts where, each part is annotated with a unique color. Similar to the keypoint based approach, as discussed in Sec. \ref{sec:proposedmethod}, we represent a semantic map as a 16 channel sparse heatmap where, each channel corresponds to the binary mask of one specific part. We use 5,161 image pairs for training and 795 image pairs for testing. The network is trained for 65K iterations keeping all other implementation conditions same as discussed in Sec. \ref{sec:proposedmethod}. Our method is able to successfully reconstruct the semantic maps by transferring visual attributes from the reference image. In Fig. \ref{fig:semantic}, we show a few qualitative results of conditional semantic reconstruction using the proposed method.

\subsubsection{Virtual Try-On}\label{sec:application2}
\begin{figure}[t]
  \centering
  \includegraphics[width=\linewidth]{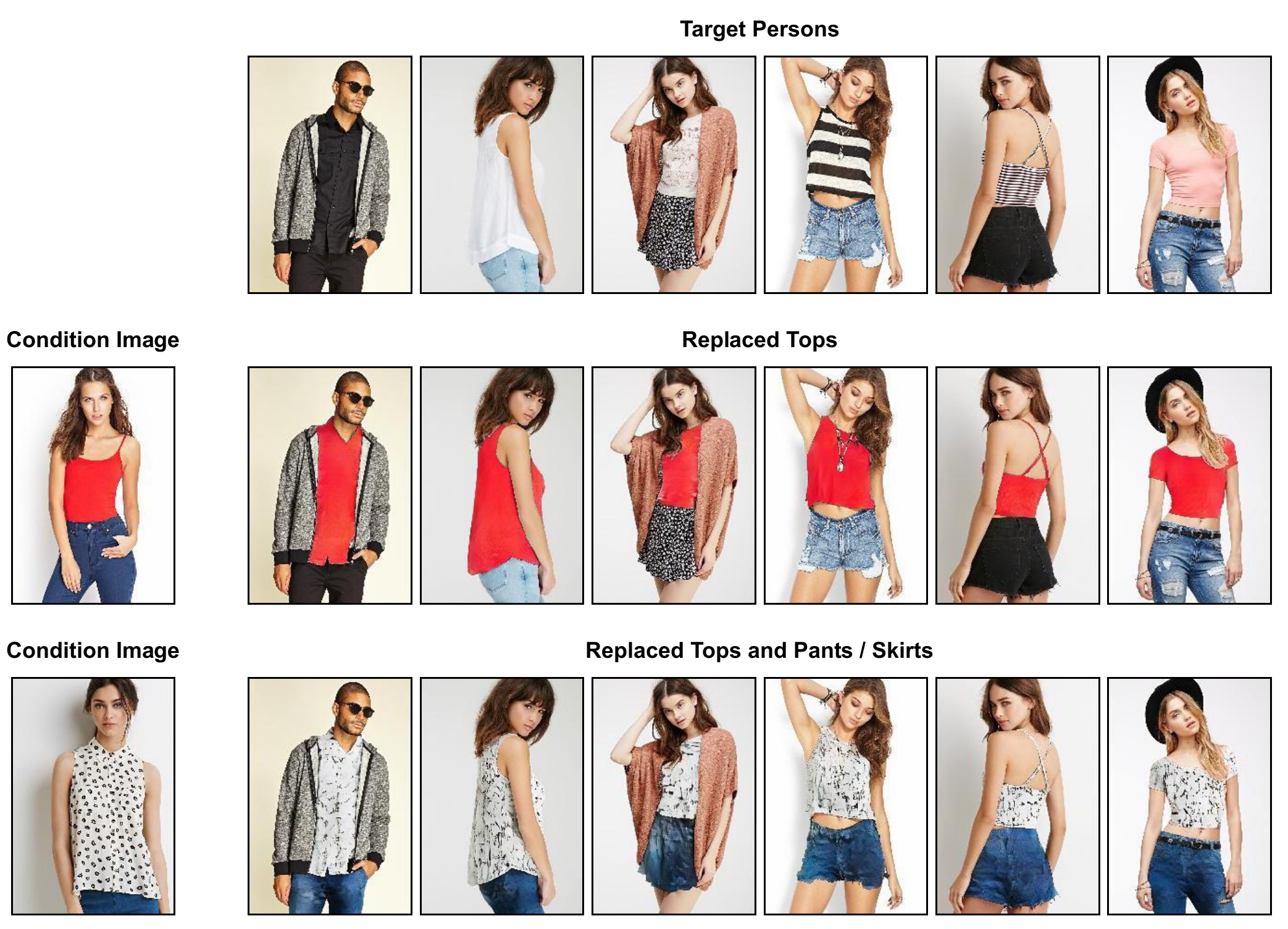}
  \caption{Qualitative results of virtual try-on using the proposed method.}
  \label{fig:viton}
\end{figure}

Although, the proposed network is not originally intended for virtual try-on applications, the ability of conditional semantic reconstruction can be further extended to address such problems to a significant extent. In this application, the main objective is to replace selected parts of the attire of a target person with that of a reference person. We achieve this in two sequential steps. Initially, we perform a crude reconstruction of the semantic map of the target person by conditioning on the reference person image. The crude reconstructed image is then refined by bitwise operations involving the target person image and the binary mask with selected parts of the attire. Mathematically,

\begin{equation*}
  I_{fine} = \left[M_{parts} \odot I_{crude}\right] \oplus \left[(1 - M_{parts}) \odot I_{target}\right]
\end{equation*}
where, $I_{fine}$ denotes the refined target person image with replaced attire, $I_{crude}$ denotes the reconstructed target person image, $I_{target}$ denotes the target person image with original attire and $M_{parts}$ denotes a binary mask of selected parts of the attire. In Fig. \ref{fig:viton}, we show a few qualitative results of virtual try-on using this approach. Our method is able to successfully replace local parts of the target person's attire.

\subsubsection{Skeleton Guided Style Transfer}\label{sec:application3}
As a part of our earlier work on a scene text editing framework, STEFANN \cite{roy2020stefann}, we propose two independent network architectures for preserving structural and color consistencies in a generative character-to-character transformation. In this cross-domain application, we show that the performance of such a two-stage style transfer technique can be significantly improved by leveraging the proposed end-to-end architecture with skeleton supervision. The structural information of the source character is provided as a single channel binary image tensor of its skeleton. However, in this case, the geometric structure of the target character is not known before inference. For this reason, we use the target character skeleton from a fixed font as a crude structural approximation of the target character. In our experiments, we train the network with a much smaller subset of the dataset of STEFANN. We randomly select 200 image pairs of uppercase characters from both training and testing sets. This leads to 203000 image pairs from 1015 fonts for training and 60000 image pairs from 300 fonts for testing, where a random color is applied for each font. The skeleton is estimated from the binarized character image by applying Gaussian blur with a $3 \times 3$ kernel followed by a parallel thinning algorithm \cite{zhang1984fast}. We train the network for 65K iterations keeping all other implementation conditions same as discussed in Sec. \ref{sec:proposedmethod}.

\begin{figure}[h]
  \centering
  \includegraphics[width=\linewidth]{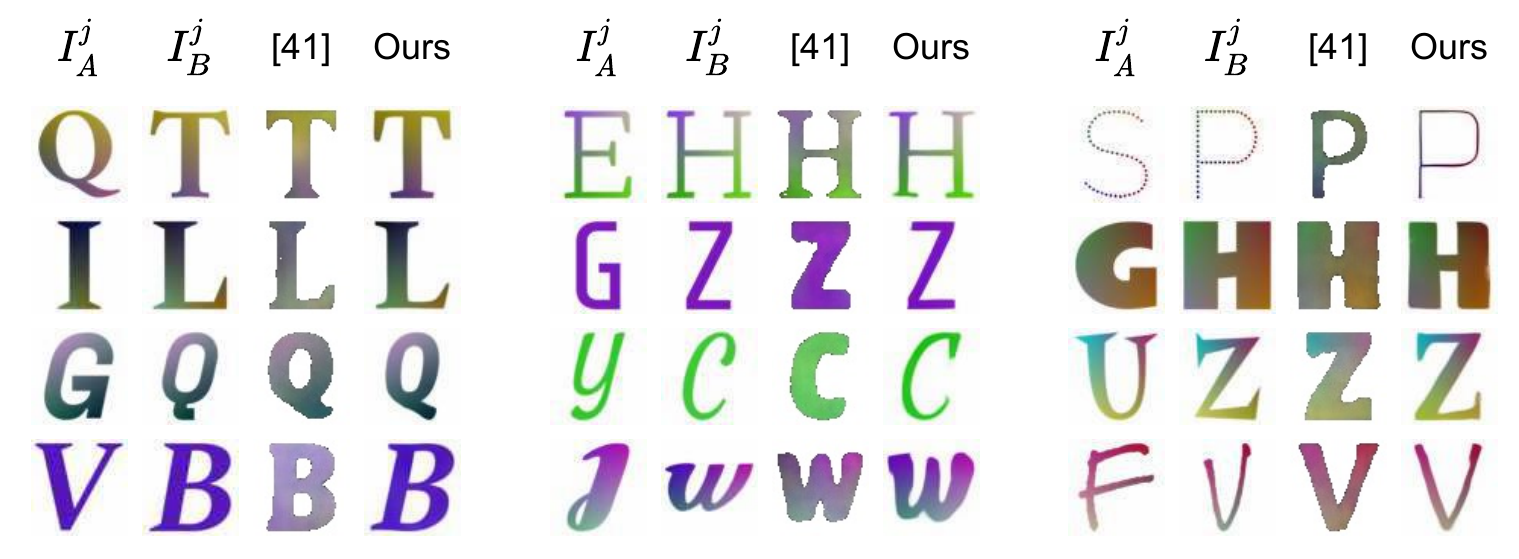}
  \caption{Qualitative comparison of different methods of style transfer. $I_A^j$ denotes the condition image, $I_B^j$ denotes the target image and subsequent columns show the generated images by STEFANN \cite{roy2020stefann} and our method using skeleton supervision.}
  \label{fig:style}
\end{figure}

\begin{table}[h]
\centering
\caption{Quantitative comparison of different methods of style transfer.}
\begin{tabular}{l|cccc}
\hline
Method & SSIM &
  \begin{tabular}[c]{@{}c@{}}PSNR\\ (dB)\end{tabular} &
  \begin{tabular}[c]{@{}c@{}}LPIPS\\ (VGG)\end{tabular} &
  \begin{tabular}[c]{@{}c@{}}LPIPS\\ (SqzNet)\end{tabular} \\ \hline
STEFANN \cite{roy2020stefann} & 0.450 & 13.347 & 0.397 & 0.273 \\
Ours                          & \textbf{0.638} & \textbf{16.876} & \textbf{0.208} & \textbf{0.089} \\ \hline
Real Data                     & 1.000 & $\infty$ & 0.000 & 0.000 \\ \hline
\end{tabular}
% \caption{Quantitative comparison of different methods of style transfer.}
\label{tab:style}
\end{table}

In Fig. \ref{fig:style}, we show a qualitative comparison of different methods of style transfer. Even with over 70\% reduction in the training data, the end-to-end skeleton guided approach can preserve structural and color consistencies significantly better than the two-stage pipeline. This visual analysis is further reflected in multiple perceptual metrics including SSIM, PSNR and LPIPS as shown in Table \ref{tab:style}.

% ---------- CONCLUSION ----------
\section{Conclusion}\label{sec:conclusion}
In this paper, we propose a keypoint based generative method for human pose transfer. The key contribution of our work is an improved network architecture by introducing attention links at every underlying resolution level in the encoding and decoding streams of the generator. Our method achieves state-of-the-art results on the DeepFashion dataset in both qualitative and quantitative benchmarks. The visual superiority of the generated images by our method is further supported by an opinion based subjective user study. We also show the efficacy of the proposed network across multiple application domains including conditional semantic reconstruction, virtual try-on and skeleton guided style transfer. These extended applications immediately encourage future works to explore domain specific modifications of the proposed end-to-end architecture.

% ---------- ACKNOWLEDGEMENTS ----------
\section*{Acknowledgments}
We would like to thank NVIDIA Corporation for providing a TITAN X GPU through the GPU Grant Program.

% ---------- BIBLIOGRAPHY ----------
\bibliographystyle{IEEEtran}
\bibliography{IEEEabrv,references}

% ---------- BIOGRAPHY ----------
% \begin{IEEEbiography}{Prasun Roy}
% Biography text.
% \end{IEEEbiography}

% \begin{IEEEbiography}{Saumik Bhattacharya}
% Biography text.
% \end{IEEEbiography}

% \begin{IEEEbiography}{Subhankar Ghosh}
% Biography text.
% \end{IEEEbiography}

% \begin{IEEEbiography}{Umapada Pal}
% Biography text.
% \end{IEEEbiography}

% ---------- APPENDIX ----------
\clearpage

\renewcommand{\thefigure}{S\arabic{figure}}
\setcounter{figure}{0}

\onecolumn

\section*{\centering \LARGE \textbf{Supplementary Material}}

\vspace{1em}

\begin{figure*}[ht]
  \centering
  \includegraphics[width=\linewidth]{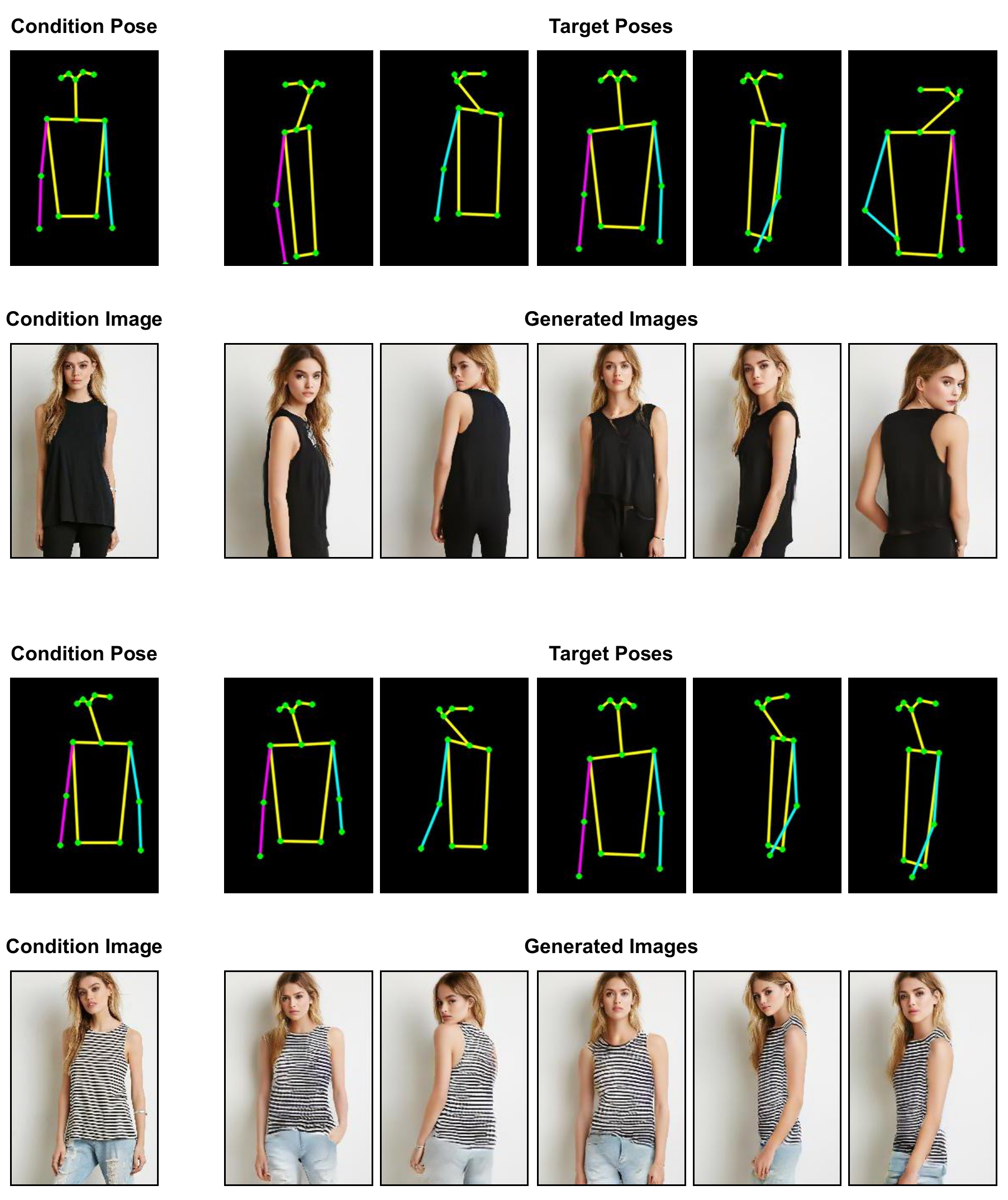}
  \caption{Qualitative results generated using the proposed method.}
  \label{fig:results1}
\end{figure*}

\begin{figure*}[ht]
  \centering
  \includegraphics[width=\linewidth]{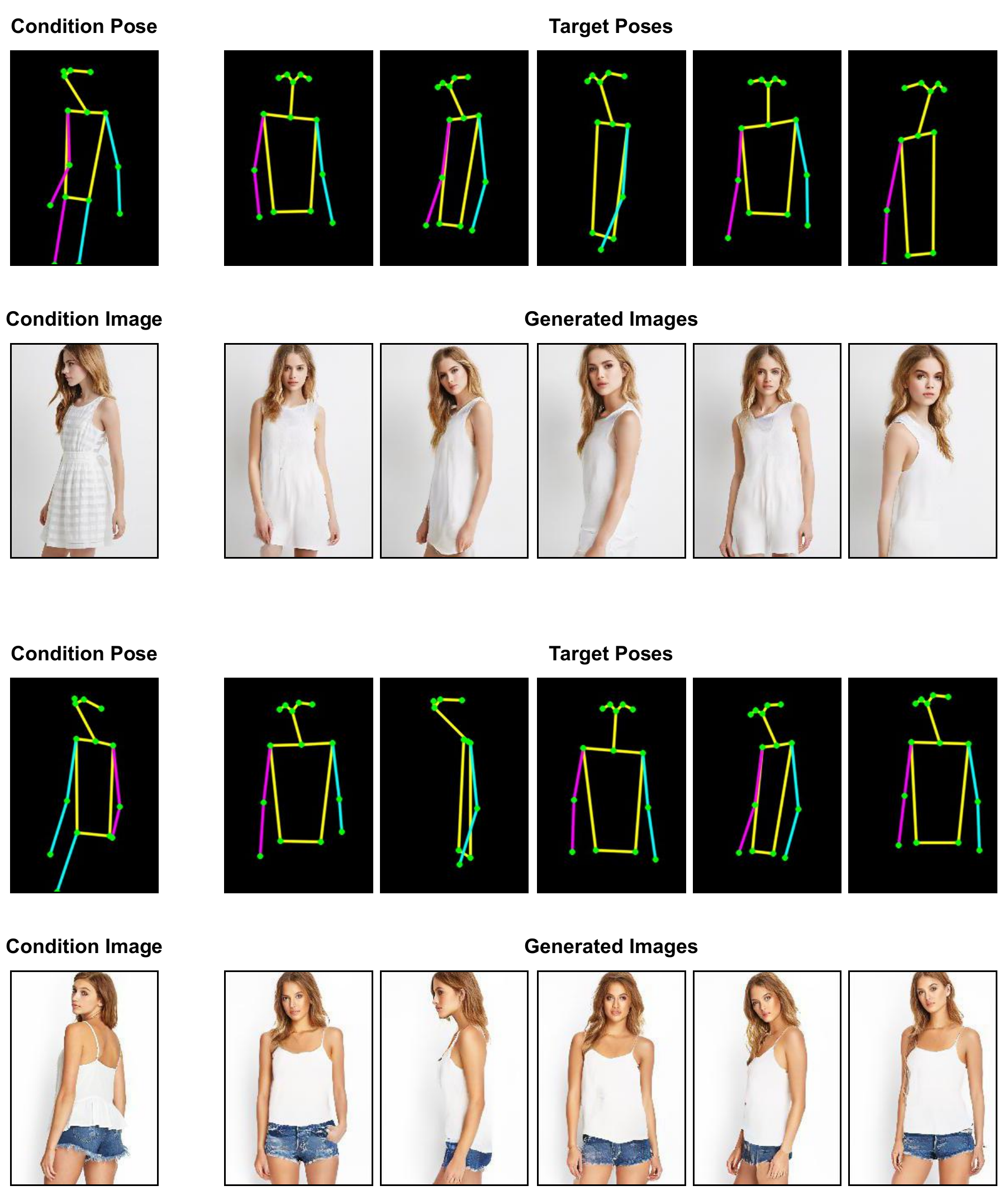}
  \caption{Qualitative results generated using the proposed method.}
  \label{fig:results2}
\end{figure*}

\begin{figure*}[ht]
  \centering
  \includegraphics[width=\linewidth]{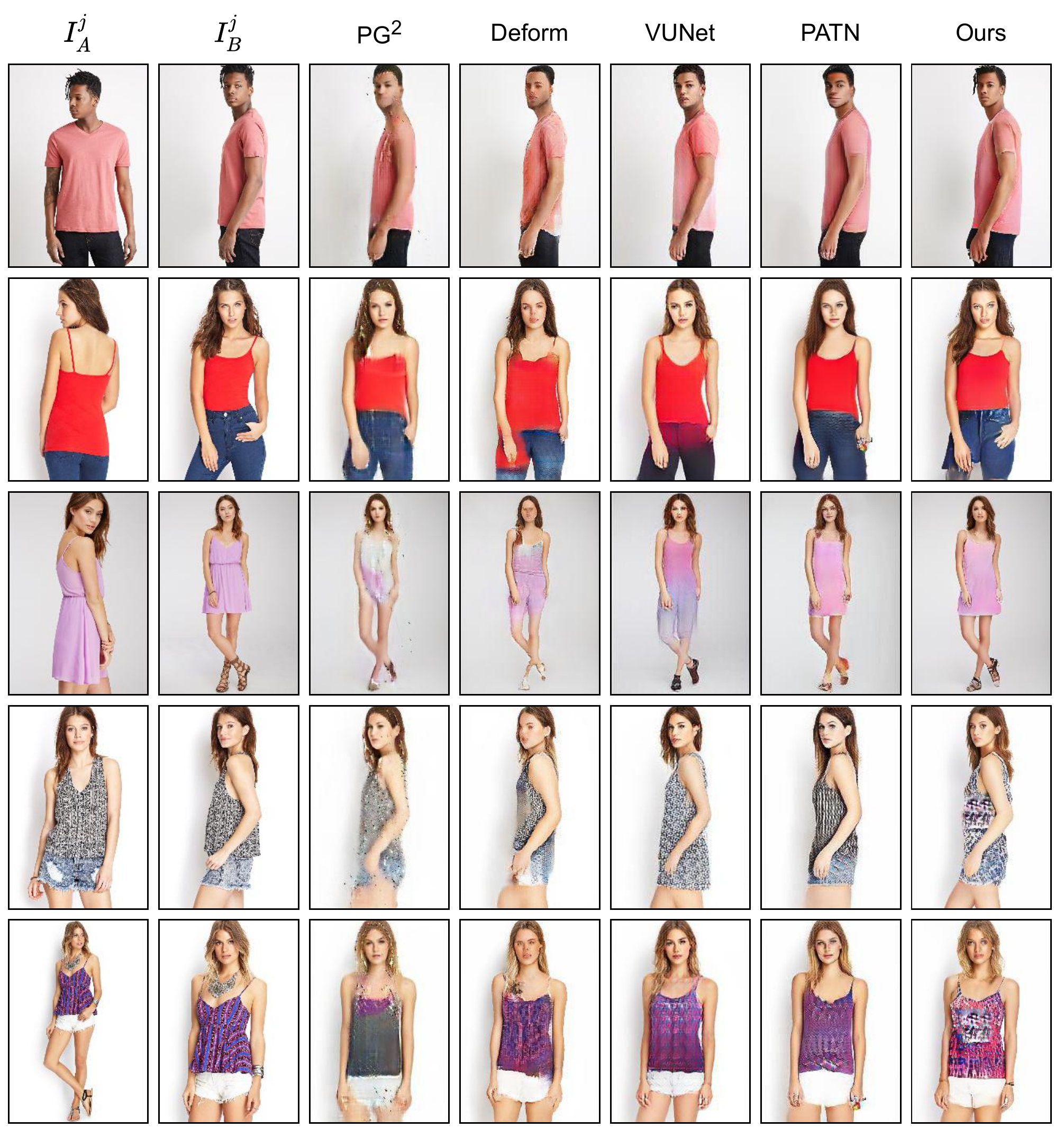}
  \caption{Qualitative comparison among different pose transfer methods. $I_A^j$ denotes the condition image, $I_B^j$ denotes the target image and subsequent columns show the generated images by $\text{PG}^2$ \cite{ma2017pose}, Deformable GANs \cite{siarohin2018deformable}, VUNet \cite{esser2018variational}, PATN \cite{zhu2019progressive} and our method.}
  \label{fig:comparison1}
\end{figure*}

\begin{figure*}[ht]
  \centering
  \includegraphics[width=\linewidth]{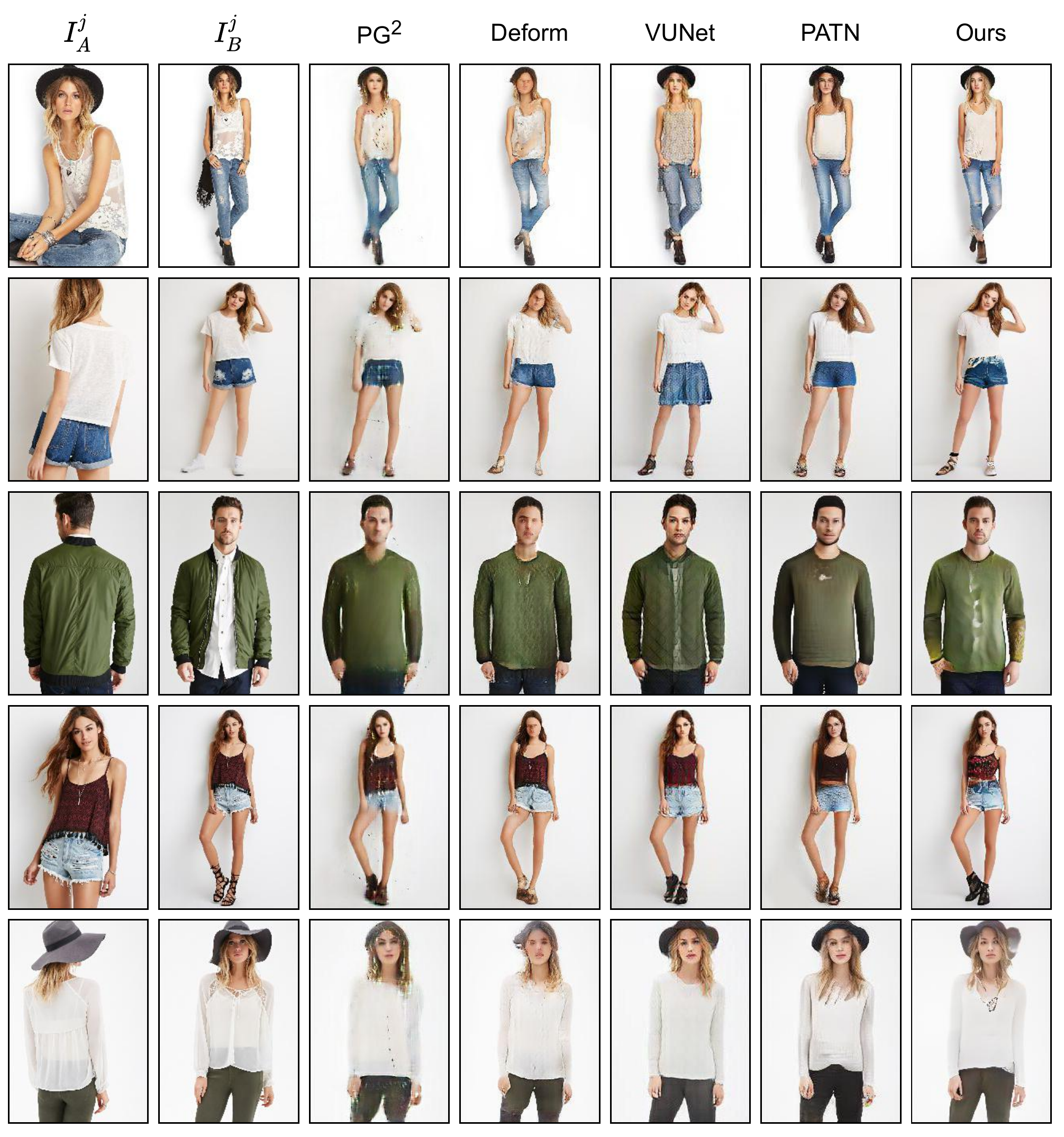}
  \caption{Qualitative comparison among different pose transfer methods. $I_A^j$ denotes the condition image, $I_B^j$ denotes the target image and subsequent columns show the generated images by $\text{PG}^2$ \cite{ma2017pose}, Deformable GANs \cite{siarohin2018deformable}, VUNet \cite{esser2018variational}, PATN \cite{zhu2019progressive} and our method.}
  \label{fig:comparison2}
\end{figure*}

In this extended text, we present more visual results as supplementary to the original paper. In Fig. \ref{fig:results1} and \ref{fig:results2}, we show visual examples of pose transfer using our method. In Fig. \ref{fig:comparison1} and \ref{fig:comparison2}, we show qualitative comparisons among previously proposed major pose transfer algorithms and our method. For a direct and fair comparison, we use the same image pairs $(I_A^j, I_B^j)$ as \cite{zhu2019progressive} and expanded their visual analysis by incorporating results of our method. Additionally, our code repository and pre-trained models are publicly available at \url{https://github.com/prasunroy/pose-transfer}.

\end{document}